\documentclass[11pt]{article}
\usepackage{graphicx} 

\usepackage{amssymb}
\usepackage{epstopdf}
\usepackage{amsmath}
\usepackage[latin1]{inputenc}

\newtheorem{theo}{Theorem}
\newtheorem{defi}{Definition}
\newtheorem{coro}{Corollary}

\title{\begin{center}On the Schoenberg Transformations\\ in Data Analysis: \\ Theory and Illustrations \end{center}}
\author{Fran\c{c}ois Bavaud\\  \\
  Department of Computer Science and Mathematical Methods\\
Department of Geography\\
University of Lausanne,  Switzerland}

\date{}

\begin{document}
\maketitle             

\begin{abstract}
The class of Schoenberg transformations, embedding Euclidean distances into higher dimensional Euclidean spaces, is presented,  and derived
 from theorems on positive definite and conditionally negative definite matrices. Original results on the arc lengths, angles and curvature of the transformations are proposed, and visualized on artificial data sets by classical multidimensional scaling. A simple distance-based discriminant algorithm illustrates the theory, intimately connected to the Gaussian  kernels of  Machine Learning.
\end{abstract}


\section{Introduction}
Schoenberg  transformations are elementwise mappings of  Euclidean distances into new 
Euclidean distances, embeddable in a higher dimensional space. Their potential  in Data Analysis seems  evident in view of the omnipresence of  Euclidean dissimilarities  in Multidimensional Scaling (MDS), Factor Analysis, Correspondence Analysis or Clustering. Yet, despite its respectable age (Schoenberg 1938a), the properties and the very existence of this class of transformations 
appear to be little   known  in  the Data Analytic community. 

Non-linear embeddings of original data into higher dimensional  feature spaces  are familiar in 
the Machine Learning community, which however bases its formalism upon  kernels, which are positive definite (p.d.) matrices, rather than on squared Euclidean distances, which are conditionally negative definite (c.n.d.)  matrices with a null diagonal. 

Some aspects of the correspondence between p.d. and c.n.d. matrices are well-known in Data Analysis, and lie at the core of classical MDS (Theorems \ref{the1} and \ref{the2}). Other aspects (Theorem  \ref{the4}),  central to  the derivation of Schoenberg transformations (Definition 
\ref{def2}), are less notorious. Section \ref{defthe} is a self-contained review of all those results, scattered in the literature, together with their proofs. 
Section \ref{spst} analyses some of the  general properties of Schoenberg transformations, and yields original  results about angles, arc lengths and curvatures. Section \ref{illustr} illustrates the non-linear and spectral properties of the transformations on two artificial data sets - the grid and the rod. An elementary yet efficient  distance-based linear discriminant algorithm is presented in Section 
\ref{adbda}. Section \ref{conc} proposes in conclusion to revisit   the  Machine Learning formalism in terms of Euclidean  distances, rather than in terms of kernels

\section{Definitions and Theorems}
\label{defthe}
\subsection{Preliminaries}
\label{prel}
Classical multidimensional scaling (MDS) (e.g. Borg and Groenen 1997) can be performed iff the eigenvalues of the so-called {\em matrix of scalar products}  are non-negative. For concision sake, we   shall refer to such a matrix as  {\em positive definite} (instead of   ``semi-positive definite"), while a
 {\em strictly 
positive definite} matrix will  be characterized by  strictly positive eigenvalues. 

Vectors are meant as column vectors. $I$ denotes the identity matrix, and ${\bf 1}$ the unit vector, all components of which being unity. Depending upon context, the ``prime" either denotes the transpose of a matrix, or the derivative of a scalar function.
 
\begin{defi}
A real symmetric $n\times n $ matrix $C=(c_{ij})$ is said to be 
\begin{itemize}
  \item[$\bullet$] {\rm  positive  definite  (p.d.)} if $(z,Cz)=\sum_{ij}^n c_{ij}z_iz_j\ge 0$ for all vectors $z\in  \mathbb{R}^n$
  \item[$\bullet$] {\rm conditionally negative  definite (c.n.d)} if $(z,Cz)=\sum_{ij}^n c_{ij}z_iz_j\le 0$
  for all $x\in  \mathbb{R}^n$ such that $\sum_{i=1}^nz_i=0$. 
 \end{itemize}
\end{defi}

Consider a {\em signed distribution} $a$ on $n$ objects, that is a vector obeying $\sum_{i=1}^n a_i=1$,
where some components are possibly negative. Consider also the  $n\times n$ {\em centering matrix} $H(a)=I-{\bf 1}a'$, with components $\delta_{ij}-a_j$. Let $C$ be a symmetric $n\times n$ matrix, and define the matrix
\begin{equation}
\label{bac}
B(a)=-\frac12\:  H(a)\:  C\:  H'(a)\enspace .
\end{equation}

\begin{theo}[Young and Householder 1938; Schoenberg 1938b]
\label{the1}
\mbox{} \\
For any signed distribution $a$, 
\begin{displaymath}
 B(a)\quad\mbox{\rm is p.d.}\qquad \Leftrightarrow\qquad C \quad\mbox{\rm is c.n.d.}
\end{displaymath}
\end{theo}

\noindent {\bf Proof:}    first observe that if $B(a)$ is p.d., then $B(\tilde{a})$ is also p.d. for any other signed distribution $\tilde{a}$,  in view of the identity
$B(\tilde{a})=H(\tilde{a})B(a)H'(\tilde{a})$, itself a consequence of $H(\tilde{a})=H(\tilde{a})H(a)$. Also, for any $z$, 
$(z,B(a)z)=-\frac12(y,Cy)$ where 
the vector
$y=H'(a)z$ obeys $\sum_i y_i=0$ for any $z$,   showing ``$\Leftarrow$".  Also, $y=H'(a)y$ whenever 
$\sum_i y_i=0$, and hence $(y,B(a)y)=-\frac12(y,Cy)$, thus 
demonstrating
``$\Rightarrow$". $\Box$

\begin{theo}[classical MDS]
\label{the2}
Let $C=(c_{ij})$ be   a symmetric $n\times n$ matrix. Define the associated {\em zero-diagonal} matrix $\hat{C}=(\hat{c}_{ij})$ as $\hat{c}_{ij}=c_{ij}-\frac12c_{ii}-\frac12c_{jj}$. Then 
\begin{equation}
\label{bac2}
B(a)=-\frac12\:  H(a)\:  \hat{C}\:  H'(a)\qquad\mbox{and}
\qquad
\hat{c}_{ij}=b_{ii}(a)+b_{jj}(a)-2b_{ij}(a)\enspace .
\end{equation}
Moreover, $C$ is  c.n.d. iff  $\hat{C}$ is  c.n.d. In this case, the components $\hat{c}_{ij}$ are 
``isometrically embeddable in $l_2$", that is  representable as squared Euclidean distances $D_{ij}$ between $n$ objects as 
\begin{equation}
\label{bac3}
\hat{c}_{ij}\equiv D_{ij}=\sum_{\alpha=1}^p(x_{i\alpha}-x_{j\alpha})^2\qquad\qquad i,j=1,\ldots,n
\end{equation}
where the object coordinates can be chosen as 
\begin{equation}
\label{coo1}
x_{i\alpha}=\sqrt{\lambda_\alpha(a)}\: u_{i\alpha}(a)
\end{equation}
where the
$\lambda_\alpha$ are the diagonal components of the diagonal matrix $\Lambda(a)$ and $u_{i\alpha}(a)$ are the components of the  orthogonal matrix $U(a)$ occurring in the spectral decomposition 
$B(a)=U(a)\Lambda(a)U'(a)$. 
 \end{theo}

\noindent {\bf Proof:}    the first identity in (\ref{bac2}) follows from $H(a){\bf 1}=0$, and the second one from $b_{ii}(a)+b_{jj}(a)-2b_{ij}(a)= c_{ij}-\frac12c_{ii}-\frac12c_{jj}$, itself a consequence of  the form (\ref{bac}) $b_{ij}(a)=
-\frac12 c_{ij}+\gamma_i+\gamma_j$ for some vector $\gamma$. The next assertion follows from 
$(y,Cy)=(y,\hat{C}y)$ whenever $\sum_iy_i=0$, and identity (\ref{bac3}) can be shown to amount to the second identity  (\ref{bac2}) by direct substitution. $\Box$

\vspace{0.1cm}

The p.d. nature of $B(a)$ (Theorem \ref{the1}) is crucial to insure the non-negativity of the eigenvalues
$\lambda_\alpha$. Identity  $H'(a)a=0$ yields $B(a)a=0$. Hence, at least one eigenvalue is zero and $p\le n-1$ in (\ref{bac3}). 

Theorems  \ref{the1} and  \ref{the2} show that any p.d. matrix $B$, or 
equivalently any 
c.n.d. matrix $C$,  define a unique set of squared Euclidean distances $D$ between objects (Torgerson 1958; Gower 1966). The latter can be shown (e.g. from (\ref{coo1})) to obey  the celebrated {\em Huygens principle}, namely
\begin{equation}
\label{huyg}
\sum_{j=1}^n a_j D_{ij}=D_{ia}+\Delta_a\qquad\qquad\qquad
\Delta_a=\frac{1}{2}\sum_{i,j=1}^n a_ia_jD_{ij}
\end{equation}
where $D_{ia}$ denotes the squared distance between object $i$ (with coordinates $x_i$) and the 
{\em $a$-barycenter}  defined by the coordinates $\bar{x}_a=\sum_j a_j x_j$.  Also, 
$\Delta_a\ge0$ interprets as  the 
average dispersion of the cloud, provided $a$ is a   non-negative distribution representing the relative weights of the objects. In the general case of a signed distribution, $\Delta_a$ is still well defined, but can be negative. 

The squared Euclidean distance between the barycenters $\bar{x}_a$ and $\bar{x}_b$ associated to two signed distributions $a$ and $b$ can also be shown to satisfy 
\begin{eqnarray}\label{shp}
 D_{ab}=-\frac12\sum_{ij}(a_i-b_i)(a_j-b_j)D_{ij} \end{eqnarray}
which directly demonstrates the c.n.d. nature of $D$ (since $z_i=a_i-b_i$ obeys $\sum_i z_i=0$). Also, 
(\ref{shp}) entails  (\ref{huyg}) with the choice $b_j=\delta_{jk}$ for some $k$.  

Substituting (\ref{huyg}) in (\ref{bac}) yields
\begin{displaymath}
b_{ij}(a)=-\frac12(D_{ij}-D_{ia}-D_{ja})
\end{displaymath}
which, by the cosine theorem,  is the matrix of the {\em scalar products} between $x_i$ and $x_j$ as measured from the origin $\bar{x}_a$. Low-dimensional factorial reconstructions (that is limiting the sum in (\ref{bac3}) to the largest eigenvalues) express a maximum amount of $\mbox{tr}(B(a))=\sum_i D_{ia}$. This quantity, without  direct interpretation,  
 is proportional to the {\em uniform} dispersion of the coordinates cloud with respect to the point $\bar{x}_a$. The dispersion $\mbox{tr}(B(a))$ is minimum when $a$ is the uniform distribution, a standard choice in classical MDS (see e.g. Mardia et al. 1979). 
 
 Concentrating the mass of $a$ on a single existing  object, typically the last one, is often proposed  for computational convenience. Other prescriptions consider 
 $a_i$ as proportional to the precision of measurement of object $i$ (see e.g. Borg and Groenen 1997), or set $a_i=0$ for objects whose behavior might   influence excessively the overall configuration, 
 as in the treatment of ``{\em supplementary elements}" in Correspondence Analysis (see e.g. Benz\'ecri 1992;  Lebart,   Morineau and  Piron 1998;  Meulman,  van der Kooij and Heiser 2004;   Greenacre and  Blasius 2006). Other choices such as the circumcenter or the incenter are discussed in Gower (1982). Note that the signed nature of $a$ allows to define an {\em external  origin} $\bar{x}_a$  lying outside the convex hull  of the $n$ points, resulting in $B_{ij}\ge0$ for all pairs. 
 
As a matter of fact, the choice of the origin $a$ and the choice of the object weights $f$ constitute two   {\em distinct} operations, as  made explicit by the following generalization of classical MDS (Cuadras and Fortiana 1996;  Bavaud 2006, 2009):

\begin{theo}[weighted MDS]
\label{the3}
Consider $n$ weighted objects with positive weights $f_i>0$ normalized to $\sum_i f_i=1$, together with a (symmetric, non-negative, zero-diagonal) pairwise dissimilarity matrix $D=(D_{ij})$. Let $ \Pi=(\pi_{ij})=\mbox{diag}(f)$, i.e. $\pi_{ij}=f_i \delta_{ij}$. 
Then $D$ is squared Euclidean iff the matrix of {\em weighted scalar products}
\begin{displaymath}
K(a)=-\frac12\:  \sqrt{\Pi}\:  H(a)\:  D\:  H'(a)\:  \sqrt{\Pi}\qquad\mbox{that is}\qquad 
K_{ij}(a)=\sqrt{f_i\:  f_j}\:   b_{ij}(a)
\end{displaymath}
is p.d. The objects coordinates can be chosen as 
\begin{equation}
\label{ouzo}
x_{i\alpha}=\mbox{\normalsize $\sqrt{\frac{\lambda_\alpha(a)}{f_i}}$}\: u_{i\alpha}(a)\qquad\mbox{with}\qquad
D_{ij}=\sum_{\alpha=1}^p(x_{i\alpha}-x_{j\alpha})^2
\end{equation}
where the eigenvalues $\lambda_\alpha(a)$ and eigenvectors $u_{i\alpha}(a)$ obtain from the spectral decomposition of 
$K(a)=U(a)\Lambda(a)U'(a)$.  Moreover, the corresponding low-dimensional factorial reconstruction, retaining  in (\ref{ouzo}) only the components $\alpha$ associated with the largest eigenvalues, express
a maximum proportion of the total inertia relatively to $a$, namely  
\begin{equation}
\label{totalvar}
\mbox{tr}(K(a))=\sum_{\alpha=1}^p\lambda_\alpha=\sum_i f_i D_{ia}=\Delta_f +D_{fa}\enspace .
\end{equation}

\end{theo}
 
\noindent The proof follows from the definitions and Theorem \ref{the2}
 by direct substitution. The last identity is a consequence of (\ref{huyg}), and shows in particular the total inertia to be minimum for $a=f$, as expected. When $f$ is uniform, the eigenvalues in Theorems \ref{the2}
 and \ref{the3} coincide  up to a factor $n$.

\subsection{The class of Schoenberg transformations}
If $A=(a_{ij}) $ and $B=(b_{ij})$ are p.d. matrices of same order, so are $cA$ for $c\ge0$, $(t_ia_{ij}t_j)$   for any vector $t$ (cf. Theorem  \ref{the3}), $A+B$, $AB$ as well as 
the element-wise  product or {\em Hadamard product}  
$A\circ B$ with components $a_{ij}b_{ij}$. The latter result (Schur theorem), can be first proved for rank-one p.d. matrices, and then extended to arbitrary ranks by matrix addition (see e.g. Horn and Johnson   1991; Bhatia 2006). Combining those facts, one obtains that the Hadamard integral power $A^{\circ p}$ 
with components $ a_{ij}^p$ (where $p\in \mathbb{N}$) or the 
Hadamard
exponential $\exp(\circ A)$ with components $\exp(a_{ij})$ are p.s.d. However, $A^{\circ \lambda}$
is generally not p.d. for $\lambda>0$, unless $\lambda\ge n-2$ (Fitzgerald  and Horn 1977). P.d. matrices $A$ such that $A^{\circ \lambda}$ is p.d. for each $\lambda\ge0$ are called {\em infinitely divisible}.

P.d. matrices are referred to as {\em kernels} in the Machine Learning community (see e.g. 
Haussler 1999; Cristianini and Shawe-Taylor 2003; Hofmann,  Sch\"olkopf and  Smola 2008;  and references therein). One of the most popular kernel is the so-called {\em radial basis function} or {\em Gaussian kernel} $\exp(-\lambda D_{ij})$.

\begin{theo}[Infinitely divisible kernels]
\label{the4}
Let $C=(c_{ij})$ be a symmetric matrix, and define $B=\exp(\circ- C)$, that is $b_{ij}=\exp(-c_{ij})$. Then 
\begin{displaymath}
\mbox{$B$ is   infinitely divisible}\qquad \Leftrightarrow\qquad \mbox{$C$ is c.n.d.}
\end{displaymath}
\end{theo}

\noindent {\bf Proof:}    (Horn and Johnson 1991 p.456):  consider 
the matrix $a_{ij}(\lambda)=(1-b_{ij}^\lambda)/\lambda$. If $B$ is infinitely divisible, 
then $(z,A(\alpha)z)\le0$ for any vector $z$ summing to zero, that is $A(\lambda)$ is c.n.d. for any $\lambda>0$. Hence $\lim_{\lambda\to 0^+}a_{ij}(\lambda)=-\ln b_{ij}$ is c.n.d., showing ``$\Rightarrow$".   Conversely, suppose $C$ is c.n.d., and define $F=-H(a)CH'(a)$ where $H(a)$ is the centering matrix of Section \ref{prel}. By Theorem \ref{the1}, $F$ is p.d., and so is $\exp(\circ F)$. But 
$\exp(f_{ij})=\exp(-c_{ij}-\eta_i-\eta_j)$ since 
$f_{ij}= -c_{ij}-\eta_i-\eta_j$ for some $\eta$. Hence 
$b_{ij}=\exp(-c_{ij})=\exp(\eta_i)\exp(f_{ij})\exp(\eta_j)$ is of the form $t_ia_{ij}t_j$ with $A$ p.d, and hence p.d. By the same reasoning, $b_{ij}^\lambda=\exp(-\lambda c_{ij})$ is p.d. for any $\lambda\ge0$, since $\lambda C$ is c.n.d. iff $C$ is c.n.d., thus proving ``$\Leftarrow$". $\Box$

\begin{coro}[Gaussian kernel]
Let $D_{ij}$ be a squared Euclidean distance. Then, for any   $\lambda\ge0$,
$\exp(-\lambda D_{ij})$ is p.d., and $\tilde{D}_{ij}(\lambda)=1-\exp(-\lambda D_{ij})$ is a squared Euclidean distance.
\end{coro}

\noindent {\bf Proof:}  the first assertion follows form Theorem \ref{the4}, and the second from Theorem \ref{the2}  together with the fact that 
$\tilde{D}_{ij}(\lambda)$ can  easily be shown to be   c.n.d. with a zero diagonal. $\Box$

\vspace{0.1cm}

More generally, any mixture of $\tilde{D}(\lambda)$ over $\lambda\ge0$ is  a squared Euclidean distance, yielding the following definition and theorem: 

\begin{defi}[Schoenberg transformations]
\label{def2}
A {\rm Schoenberg transformation} is a function 
 $\varphi(D)$ from $\mathbb{R}^+$ to $\mathbb{R}^+$  of the form (Schoenberg 1938a) 
\begin{equation}
\label{schtr}
\varphi(D)=\int_0^\infty \frac{1-\exp(-\lambda D)}{\lambda}\: g(\lambda)\: d\lambda
\end{equation}
where $g(\lambda) \:  d\lambda$ is a non-negative measure on $[0,\infty)$ such that $\int_{1}^\infty \frac{g(\lambda)}{\lambda}  d\lambda<\infty$. 
\end{defi}
Note that (\ref{schtr}) entails $\varphi(D)\ge 0$ and $\varphi(0)=0$ together with 
\begin{equation}
\label{schtrprime}
\varphi'(D)=\int_0^\infty  \exp(-\lambda D) \: g(\lambda)\: d\lambda
\end{equation}
where $\varphi'(D)$ denotes the {\em derivative} of $\varphi(D)$. 

\begin{theo}[Fundamental property of  Schoenberg transformations]
\label{the5}
Let $D$ be a $n\times n$ matrix of squared Euclidean distances. Define the components of the  $n\times n$  matrix $\tilde{D}$ as 
 $\tilde{D}_{ij}=\varphi(D_{ij})$, where $\varphi(D)$ is a Schoenberg transformation. Then $\tilde{D}$ is a squared Euclidean distance.
\end{theo}

It follows from above that  all componentwise  transformations of the form $\tilde{D}_{ij}=\varphi(D_{ij})$ transform a squared Euclidean distance into another squared Euclidean distance. In his paper (1938a), Schoenberg indeed proved (Theorem 6 p. 828) that {\em all} such transformations are given by Definition \ref{def2}. More precisely, Schoenberg addressed and solved  the question of determining the class $\Phi_m$ of all the transformations $\tilde{D}=\varphi(D)$  of squared Euclidean distances $D$, associated to any configuration in $\mathbb{R}^p$, which are isometrically embeddable in an Euclidean space of sufficiently large dimensionality, that is in an Hilbert space $\mathbb{R}^{\infty}$. By construction, 
$\Phi_1 \supset \Phi_2\supset\ldots\supset\Phi_\infty$, and Definition \ref{def2} characterizes the class
$\Phi_\infty=\cap_{p\ge1}\Phi_p$. The class $\Phi_1$ is central to Brownian and fractional Brownian motion (see e.g. Alpay et al. 2009), while lower-order classes 
$\Phi_{p\le 3}$  are fundamental in 
Geostatistics (see e.g. Christakos 1984) and spatial interpolation  (see e.g. Micchelli 1986; Stein 1999). 

 \section{Some  properties of the  Schoenberg transformations}
 \label{spst}
 \subsection{Complete monotonicity}
By construction, $\varphi'(D)$ in (\ref{schtrprime}) coincides with the class of {\em completely monotonic functions} 
$f(D)$ obeying $(-1)^n f^{(n)}(D)\ge0$ (Bernstein 1929). Hence Schoenberg transformations are characterized by $\varphi(D)\ge0$ with $\varphi(0)=0$, with positive odd derivatives $\varphi'(D)$, $\varphi'''(D)$, etc., and negative even derivatives $\varphi''(D)$, $\varphi''''(D)$, etc. (see Table 1). 

\begin{displaymath}
\hspace{-0.3cm}\begin{array}{|ll|l|c|c|}\hline
\qquad \mbox{{function $g(\lambda)$}}   &   &\qquad{\mbox{transformation $\varphi(D)$}} & \mbox{bounded} & \mbox{rectifiable}  \\
\hline
g_1(\lambda)=\delta(\lambda-a)& \quad a\ge0   &  \varphi_1(D)=\frac{1-\exp(-a D)}{a} & \checkmark & \checkmark \\
g_2(\lambda)=\mbox{\small $\theta(\lambda\le\frac{\pi}{2})\: \lambda\: \sin\lambda$}    &   &  \varphi_2(D)=\frac{D(D+\exp(-\frac{\pi}{2}D))}{1+D^2} & \checkmark & \checkmark \\
g_3(\lambda)=\exp(-a\lambda) & \quad a>0   &  \varphi_3(D)=\ln(1+\frac{D}{a}) & - & \checkmark\\
g_4(\lambda)=\lambda\exp(-a\lambda)& \quad a>0   &  \varphi_4(D)=\frac{D}{a(a+D)}  & \checkmark & \checkmark \\
g_5(\lambda)=\frac{a}{\Gamma(1-a)}\lambda^{-a}&   \mbox{\scriptsize $0<a<1$}   &  \varphi_5(D)=D^a & - & - \\
\mbox{\small see Berg et al. (2008)} &      &  \varphi_6(D)=\frac{D^a}{1+D^a} \quad \mbox{\small $0<a<1$} & \checkmark & - \\\hline       \end{array}
\end{displaymath}
\begin{center} {\em Table 1: some Schoenberg transformations}\end{center}

In particular, $\sqrt{D}$ is Euclidean whenever $D$ is Euclidean. 
Also, the identity transformation $\varphi(D)=D$ obtains from $g(\lambda)=\delta(\lambda)$. The latter contribution can be made explicit in the following variant,  equivalent to Definition \ref{def2} (see e.g. Berg et al.  2008): 
\begin{displaymath}
\varphi(D)=b\: D+
\int_0^\infty (1-\exp(-\lambda D))\: d\mu(\lambda)
\end{displaymath}
where $\mu$ is a non-negative measure on $(0,\infty)$ such that $\int_0^\infty\frac{\lambda}{1+\lambda}\: d\mu(\lambda)<~\infty$ and $b\ge0$. 

There exists an important literature about {\em Bernstein functions} (see e.g. Berg et al.  2008;  Schilling et al.  2010; and references therein), defined as the  smooth non-negative functions whose first derivatives  are completely monotonic. Hence, Schoenberg transformations coincide with the class of   Bernstein functions which are zero at the origin, in the same way that Euclidean distances  are 
c.n.d matrices with zero diagonal  (Theorem \ref{the2}). 

By construction, Schoenberg transformations are closed under composition, as exemplified by 
  $\varphi_6= \varphi_4\circ \varphi_5$ in Table 1.

\subsection{Arc length; rectifiable and bounded transformations}
A Schoenberg transformation acts as an anamorphosis between Euclidean spaces: to any initial configuration of points $X$, with  mutual  squared Euclidean distances $D(X)$, corresponds a transformed configuration $\tilde{X}$ reconstructible by MDS from $\tilde{D}=\phi(D)$. By construction, the mapping 
$\tilde{X}(X)$ is unique up to a translation and a rotation.

Consider a smooth curve $C$ whose arc length is parameterized by $s$, containing two close points at mutual distance $\Delta s$. The corresponding distance on the transformed curve $\tilde{C}$ is $\Delta \tilde{s}=\sqrt{\varphi((\Delta s)^2)}$. By l'Hospital's rule, the ratio of the infinitesimal arc lengths is
\begin{displaymath}
\frac{d\tilde{s}}{ds}=\lim_{\Delta s\to 0}\frac{\sqrt{\varphi((\Delta s)^2)}}{\Delta s}=\sqrt{\varphi'(0)}
\end{displaymath}
which might  be finite or not. On the other hand, infinitely distant points in the original space might  be infinitely distant or not in  the transformed space: 
\begin{defi}
The transformation $\varphi(D)$ is said to be 
\begin{enumerate}
  \item[$\bullet$] {\rm rectifiable} if $\varphi'(0)<\infty$, that is iff  $\int_0^\infty   g(\lambda)  \: d\lambda<\infty$
  \item[$\bullet$] {\rm bounded} if $\varphi(\infty)<\infty$, that is iff  $\int_0^\infty  \frac{g(\lambda)}{\lambda} \: d\lambda<\infty$. 
  \end{enumerate}
\end{defi}

\subsection{Right angles}
Consider a triangle $ijk$ with a right angle in $k$. Hence $D_{ij}=D_{ik}+D_{jk}$  by Pythagoras' theorem. Yet, in the transformed space, $\tilde{D}_{ij}\le \tilde{D}_{ik}+\tilde{D}_{jk}$ since  $\varphi(D_1+D_2)\le \varphi(D_1)+\varphi(D_2)$, which  can be demonstrated by integrating $(1-\exp(-\lambda D_1))(1-\exp(-\lambda D_2))\ge0$ as in (\ref{schtr}). That is, {\em  the Schoenberg transformation $\tilde{\alpha}$ of a right angle $\alpha=\pi/2$ is in general acute}.  By the cosine theorem,  
\begin{equation}
\label{csolkjh}
\cos\tilde{\alpha}=\frac{\varphi(D_1)+\varphi(D_2)-\varphi(D_1+D_2)}{2\sqrt{\varphi(D_1)\varphi(D_2)}}\ge0\enspace . 
\end{equation}
Under uniform  linear dilatation of the original  right-angled  triangle by a factor $\epsilon>0$, (\ref{csolkjh}) readily yields that $\lim_{\epsilon\to\infty}\tilde{\alpha}(\epsilon)=\pi/3$ whenever $\varphi$ is bounded,
and  $\lim_{\epsilon\to0}\tilde{\alpha}(\epsilon)=\pi/2$ whenever $\varphi$ is rectifiable.

\subsection{Curvature}
\label{curv}
Straight lines are bent by 
Schoenberg transformations: think of a rod whose linear distances $d$ between constituents are contracted as, say, $\sqrt{d}$. The curvature in the transformed space can be measured as follows: consider in the original space three aligned points $i,k,j$ with $d_{ik}=d_{kj}=\epsilon$ and $d_{ij}=2\epsilon$. The {\em Menger's curvature} $\kappa$  is defined as the limit (Blumenthal 1953 p. 75) 
\begin{displaymath}
\kappa=\lim_{\epsilon\to0}\frac{4 \tilde{A}_{ijk}(\epsilon)}{\tilde{d}_{ij}(\epsilon)\: \tilde{d}_{jk}(\epsilon)\: \tilde{d}_{ik}(\epsilon)}
\end{displaymath}
where $\tilde{A}_{ijk}$ is the area of the triangle $ijk$ in the transformed space and $\tilde{d}$ denotes the length of the corresponding sides. Heron's  formula 
\begin{displaymath}
16\:  \tilde{A}^2_{ijk}=(\tilde{d}_{ij}+\tilde{d}_{jk}+\tilde{d}_{ki})
(-\tilde{d}_{ij}+\tilde{d}_{jk}+\tilde{d}_{ki})
(\tilde{d}_{ij}-\tilde{d}_{jk}+\tilde{d}_{ki})
(\tilde{d}_{ij}+\tilde{d}_{jk}-\tilde{d}_{ki})
\end{displaymath}
yields after simplification
\begin{displaymath}
\kappa^2= \lim_{\epsilon\to0}\frac{4\varphi(\epsilon^2)-\varphi(4\epsilon^2)}{\varphi^2(\epsilon^2)}=
- \frac{6\: \varphi''(0)}{(\varphi'(0))^2}\ge 0 
\end{displaymath}
where  l'Hospital's rule has been used twice  in  the last equality, under the assumption of   rectifiability.

\section{Illustrations}
\label{illustr}
\subsection{Grid}
\begin{figure}
\begin{center}
\includegraphics[width=5cm]{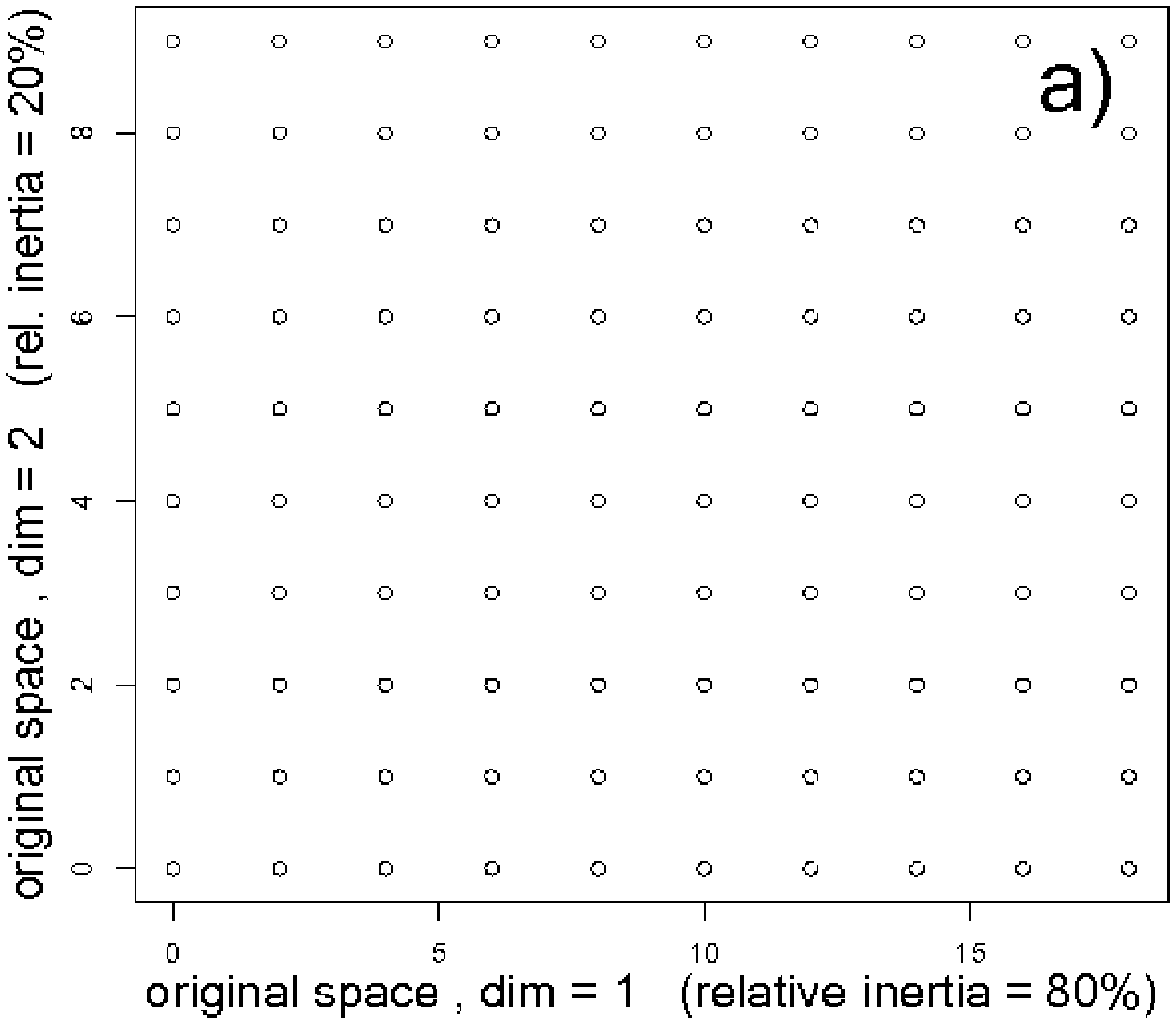}\hspace{1cm}
\includegraphics[width=5cm]{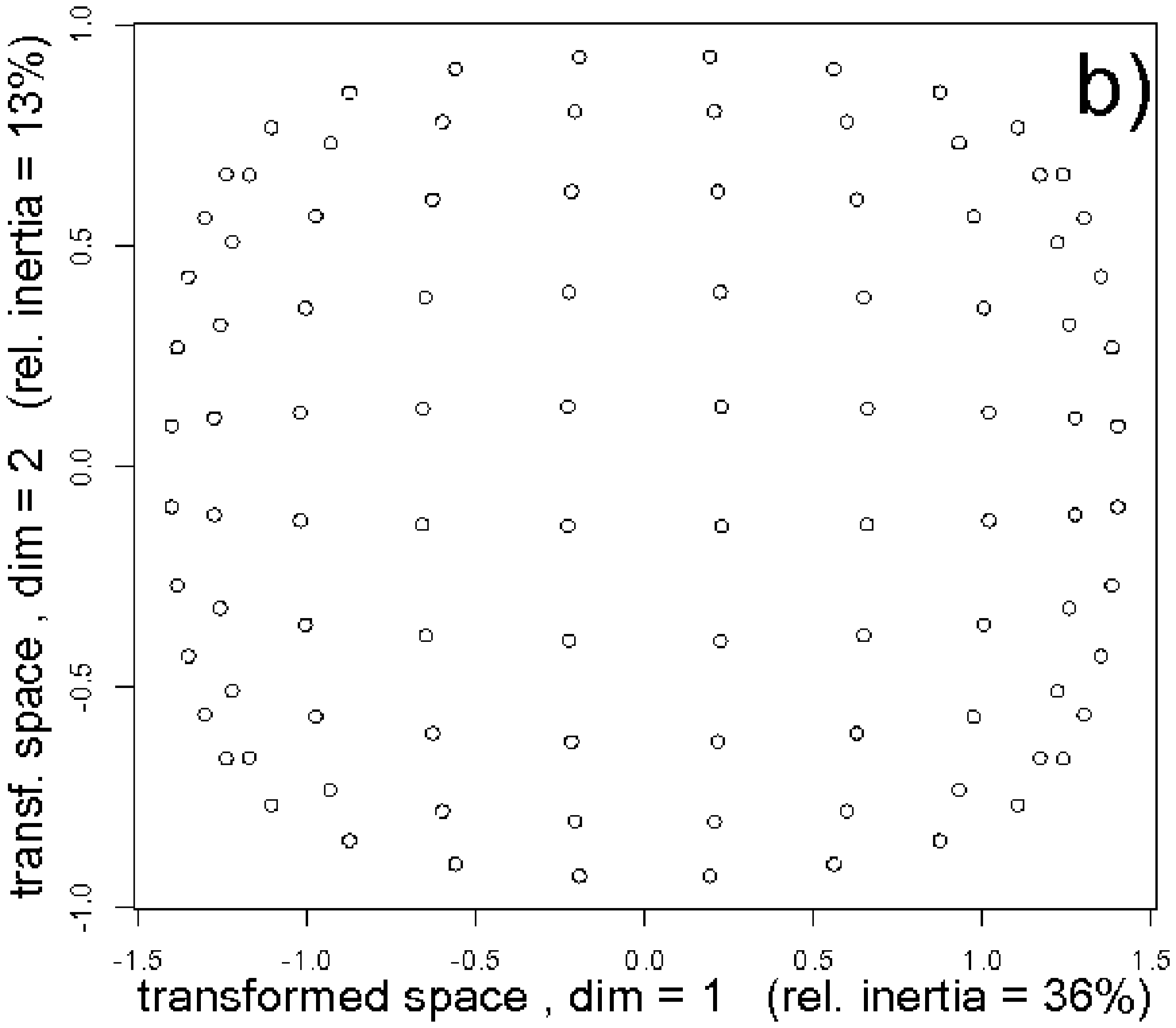}
\includegraphics[width=5cm]{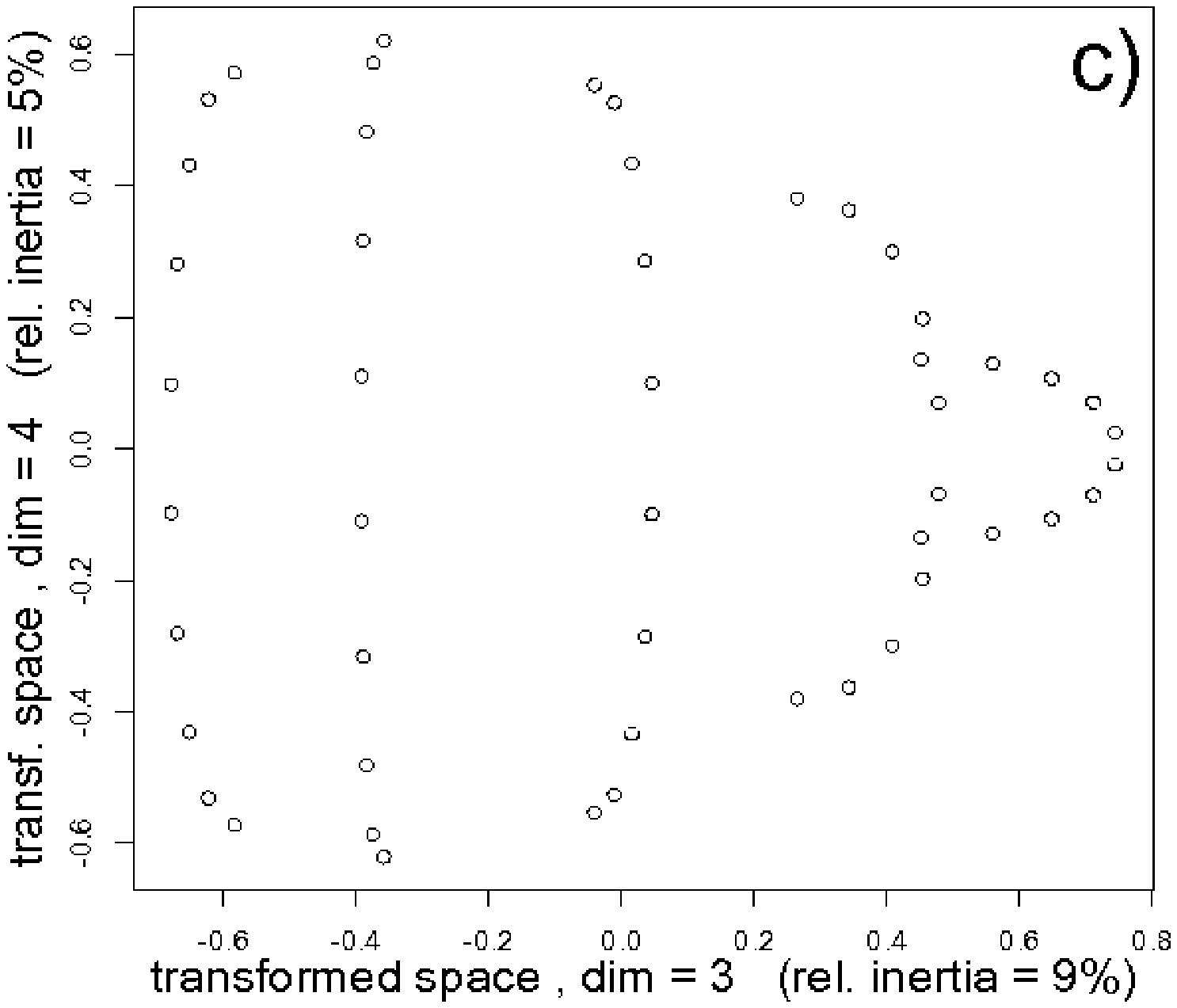}\hspace{1cm}
\includegraphics[width=5cm]{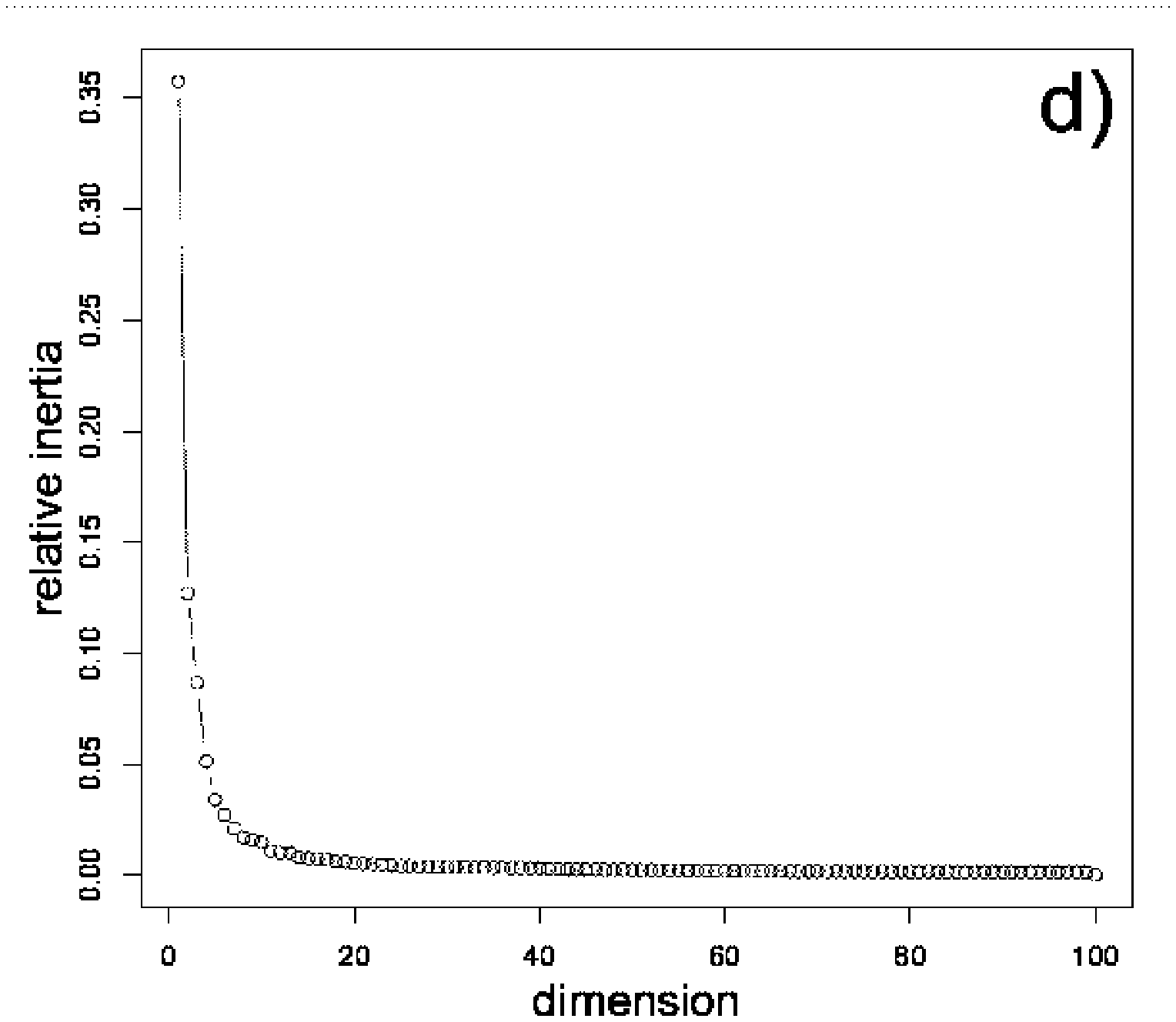}
\caption{a) Initial configuration, on which the transformation $\varphi(D)=D^{0.4}$ is applied. b) and c) depict the low-dimensional reconstruction of the transformed configuration, obtained  by weighted MDS
(Theorem \ref{the4}) where $a=f$ is the uniform distribution. d) Scree graph, proportional to the eigenvalues (\ref{totalvar}).}
\label{grid}
\end{center}
\end{figure}
Consider $n=100$ points forming the bidimensional grid of Figure \ref{grid}a), on which the transformation $\varphi(D)=D^{0.4}$ is applied. Figures \ref{grid}b) and \ref{grid}c) depict  the four first dimensions of the transformed configuration, expressing altogether 62\% of the total inertia.

\subsection{Rod}
\begin{figure}
\begin{center}
\includegraphics[width=5cm]{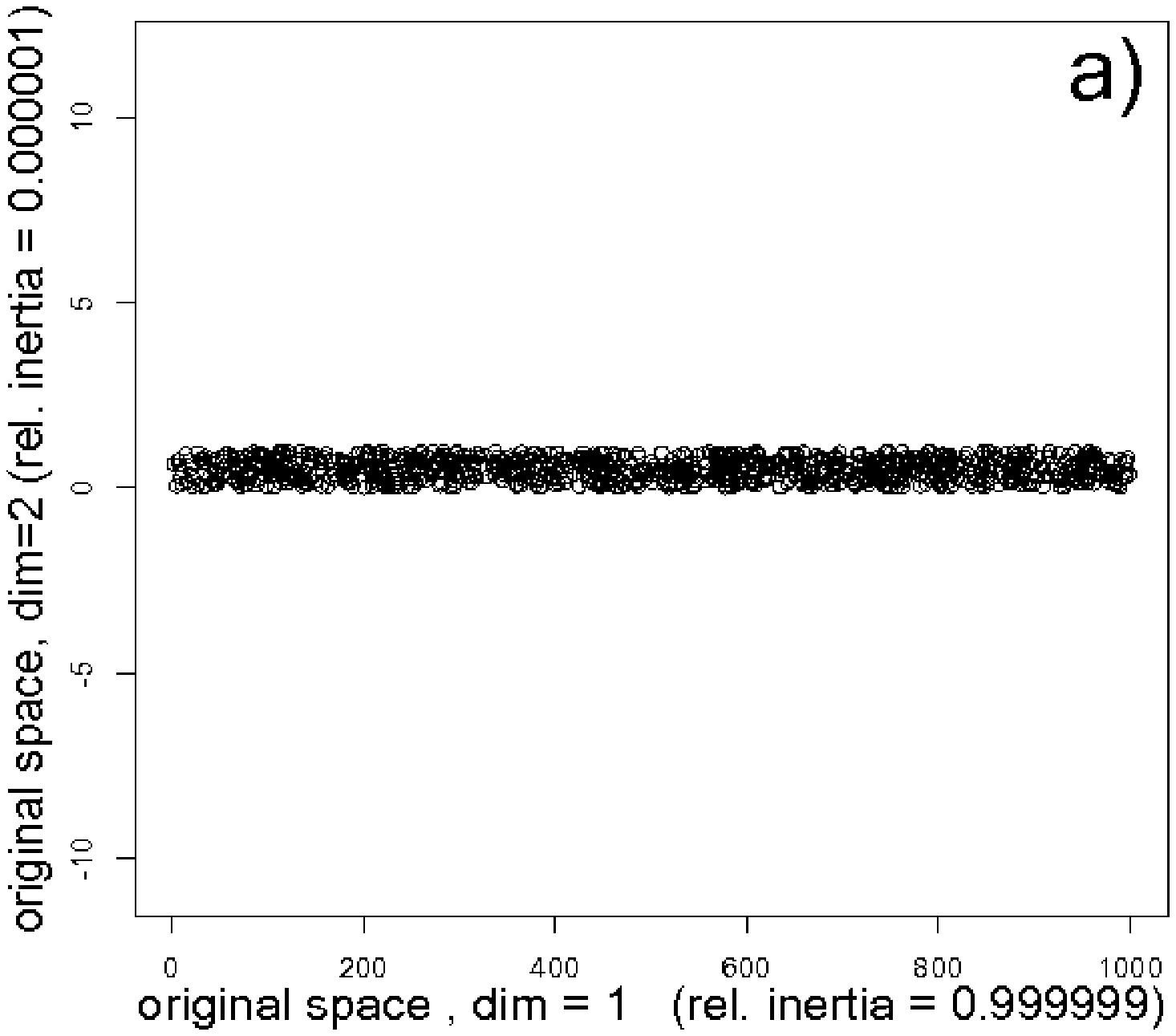}\hspace{1cm}
\includegraphics[width=5cm]{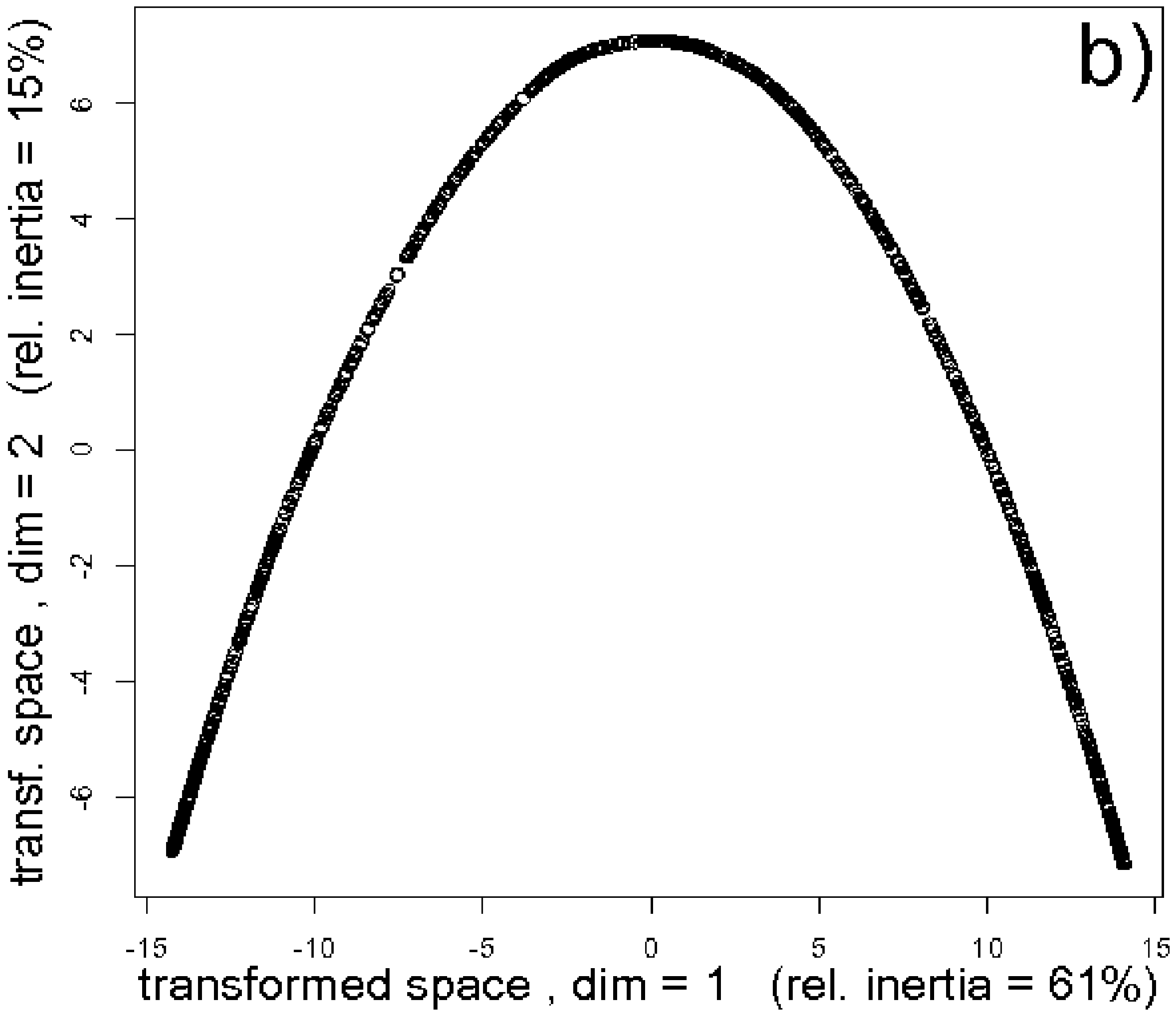}
\includegraphics[width=5cm]{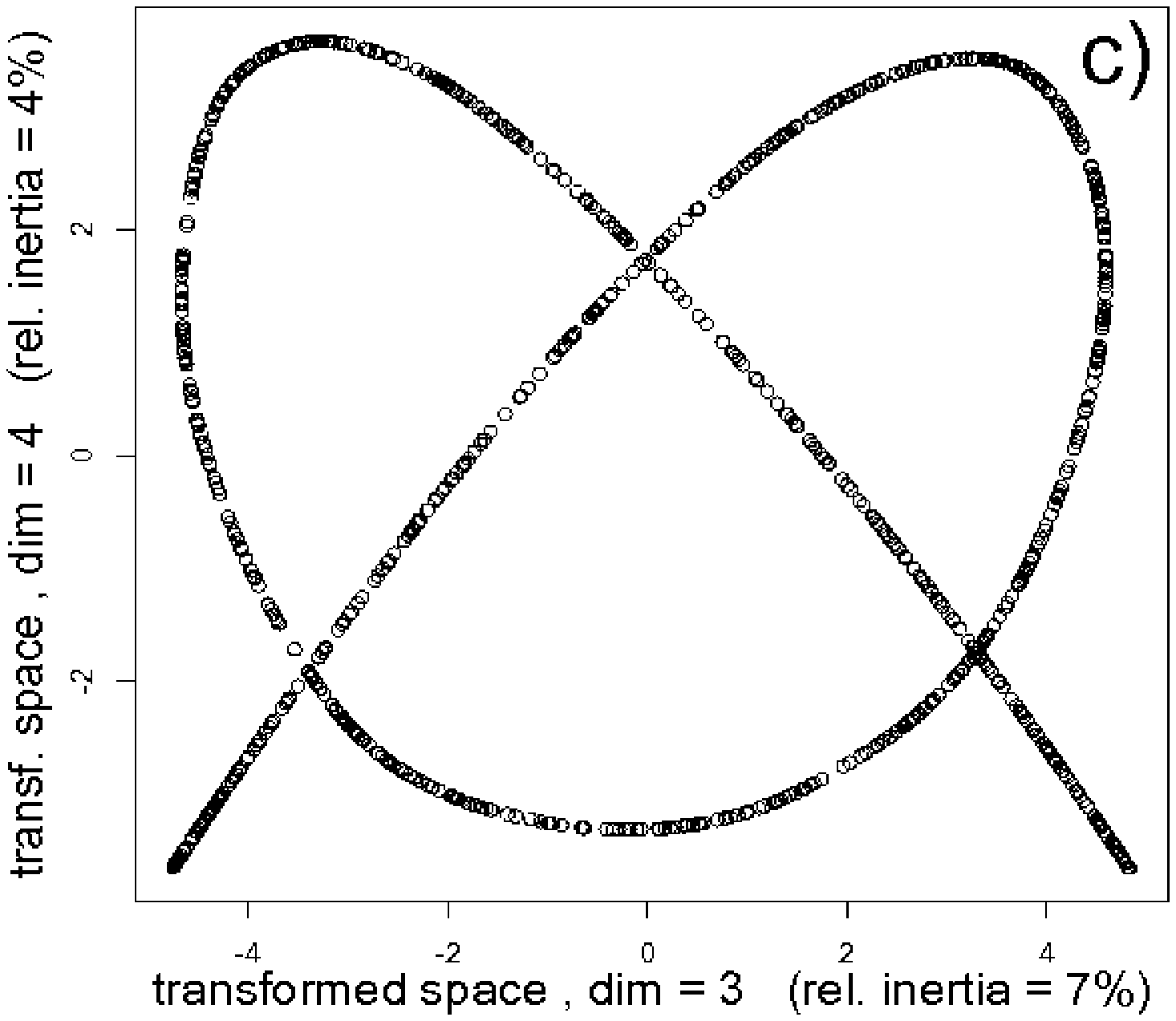}\hspace{1cm}
\includegraphics[width=5cm]{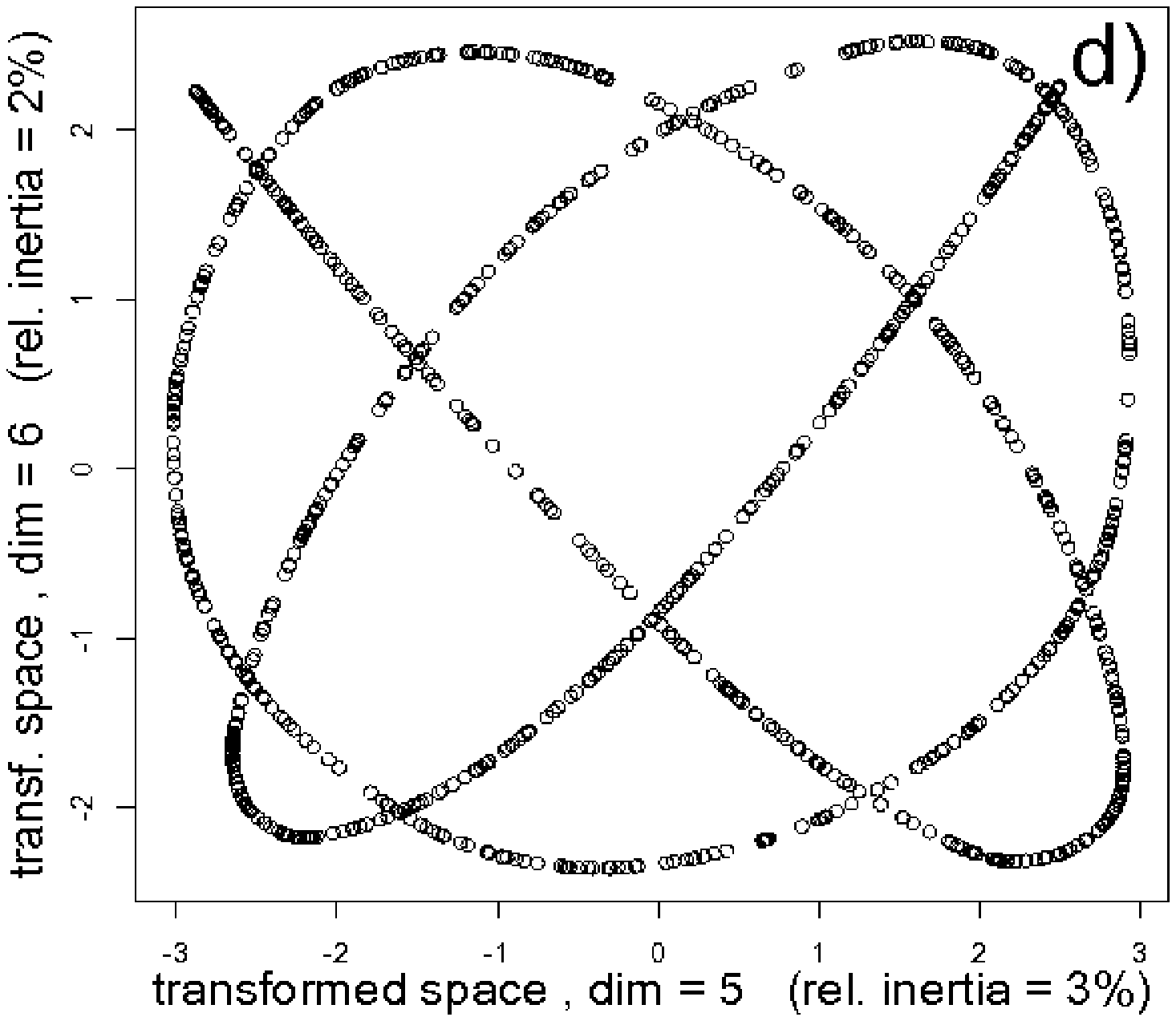}
\includegraphics[width=5cm]{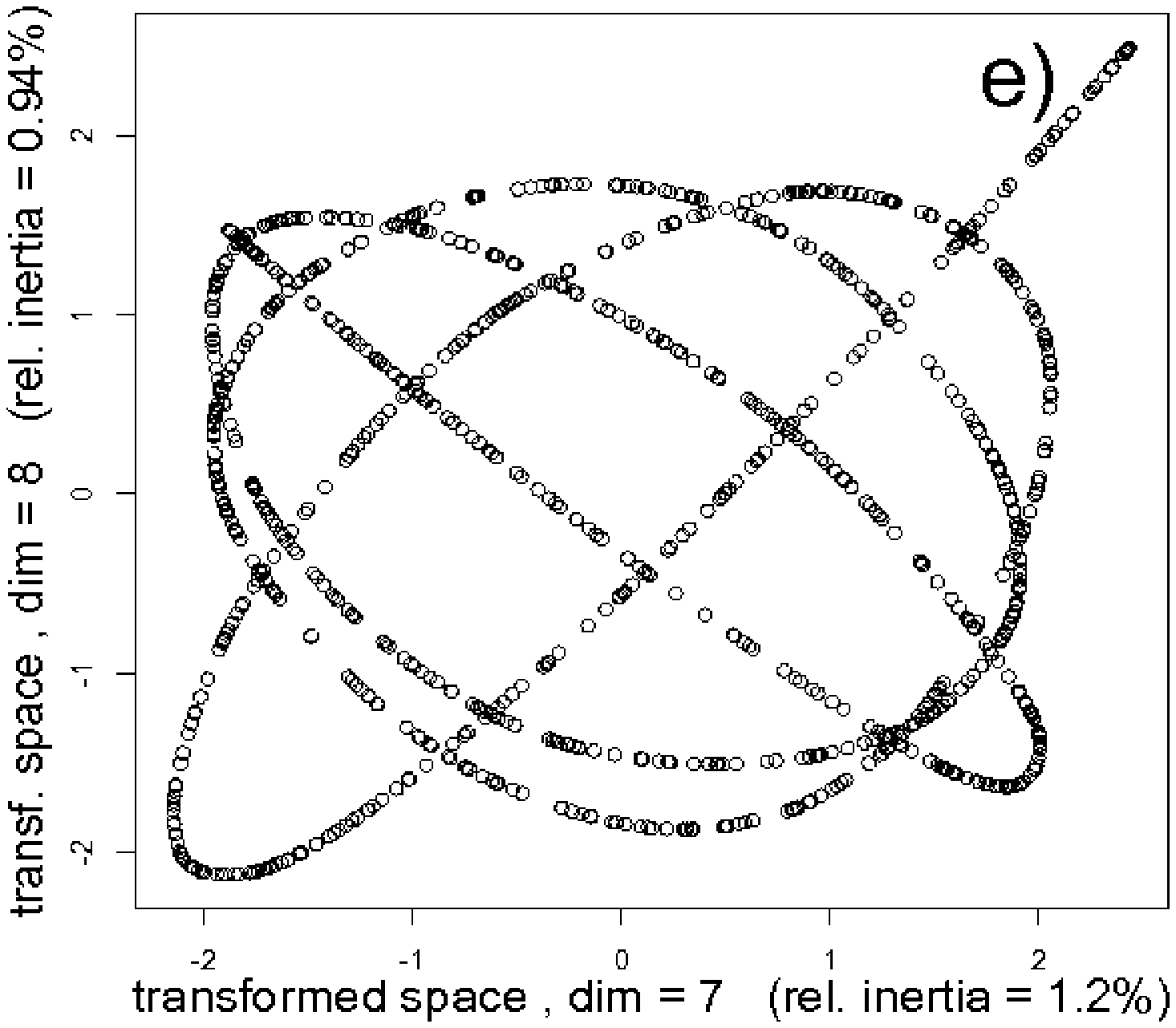}\hspace{1cm}
\includegraphics[width=5cm]{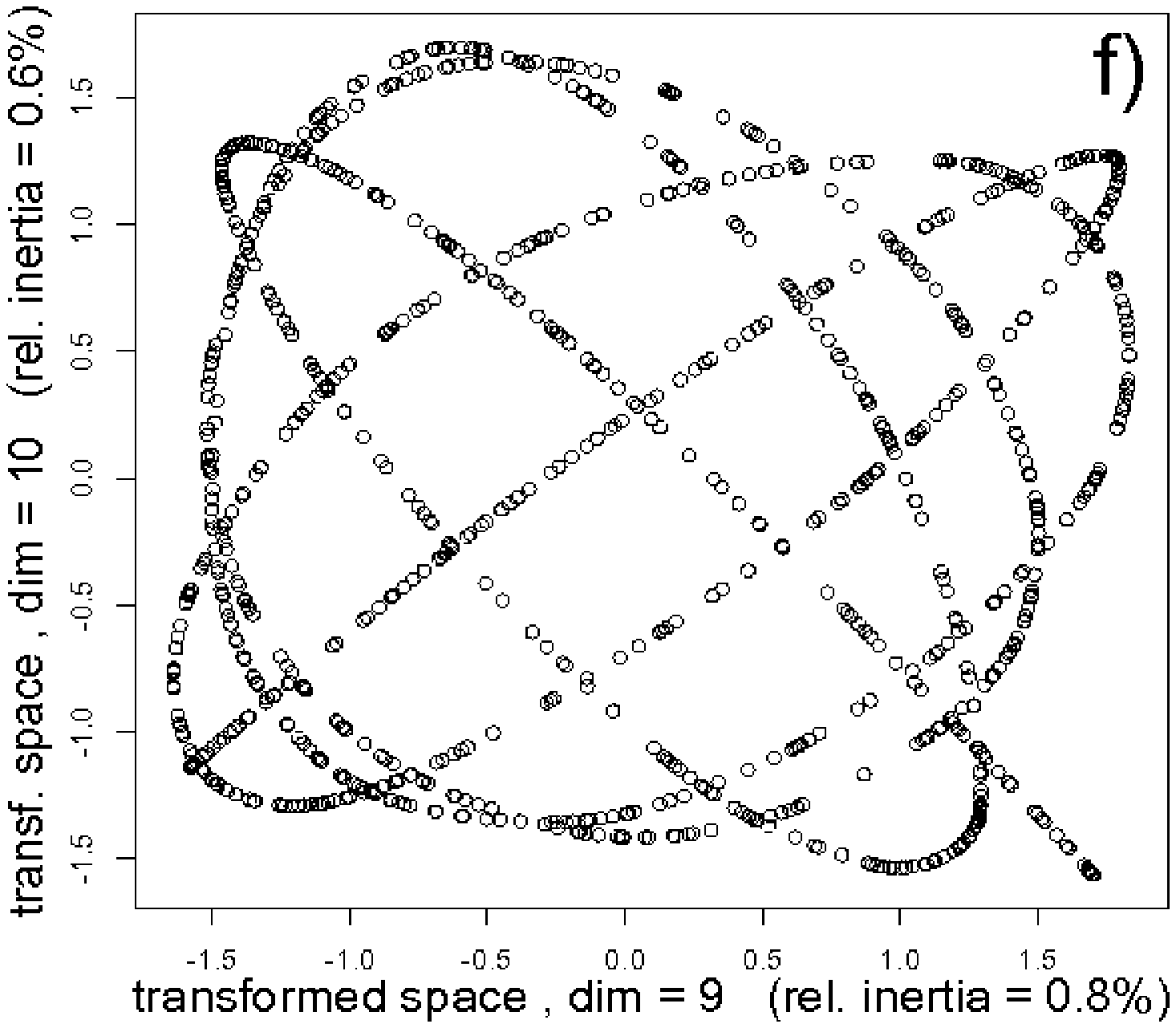}
\caption{Low-order projections (b, c, d, e and f) of the square root transformation $\tilde{D}=\sqrt{D}$ of a finite rod (a).}
\label{rod}
\end{center}
\end{figure}
Figure \ref{rod} depicts the low-order projections (b, c, d, e and f) of the non-rectifiable square root transformation $\tilde{D}=\sqrt{D}$ of a quasi-unidimensional rod of $n=1'000$ points,  uniformly generated as $X_1\sim U(0,1000)$ and $X_2\sim U(0,1)$ (a).
 As expected, the transformed rod is bent, although the  curvature formula of Section \ref{curv} does not applies here $(\varphi'(0)=\infty$). 
 
The transformation of a line is called  ``screw line" by Von Neumann and Schoenberg (1941), and ``helix" by Kolmogorov (1940) -  an adequate terminology in view of Figure \ref{rod}. 

The first MDS dimensions turn out to express 61.0\%, respectively  15.1\% of the relative inertia. Analytic arguments, to be developed in a forthcoming publication, demonstrate the corresponding exact quantities to be $\frac{6}{\pi^2}=60.8\%$, respectively $\frac{15}{2\pi^2}=15.2\%$ for a line.

\section{Application: distance-based discriminant analysis}
\label{adbda}
Consider a collection of  objects $i=1,\ldots,n$ endowed with $p$-dimensional features, yielding 
squared Euclidean distances $D_{ij}$ between  objects, possibly after standardization and/or orthogonalization of the features (Mahalanobis distances). Also, suppose that each object belongs to a group 
$g=1,\ldots m$. 
An elementary discriminant strategy would consist in assigning each object $i$ to the group $g$ whose centroid is the closest to $i$, that is to assign $i$ to $\arg\min_g D_{ig}$: this
is the linear discriminant prescription of  Fisher (1936), successfully applied on the Iris Data ($n=150$, 
$p=4$, $m=3$) with a percentage  of well-classified individuals as high as 97\%.

\begin{figure}
\begin{center}
\includegraphics[width=3.8cm]{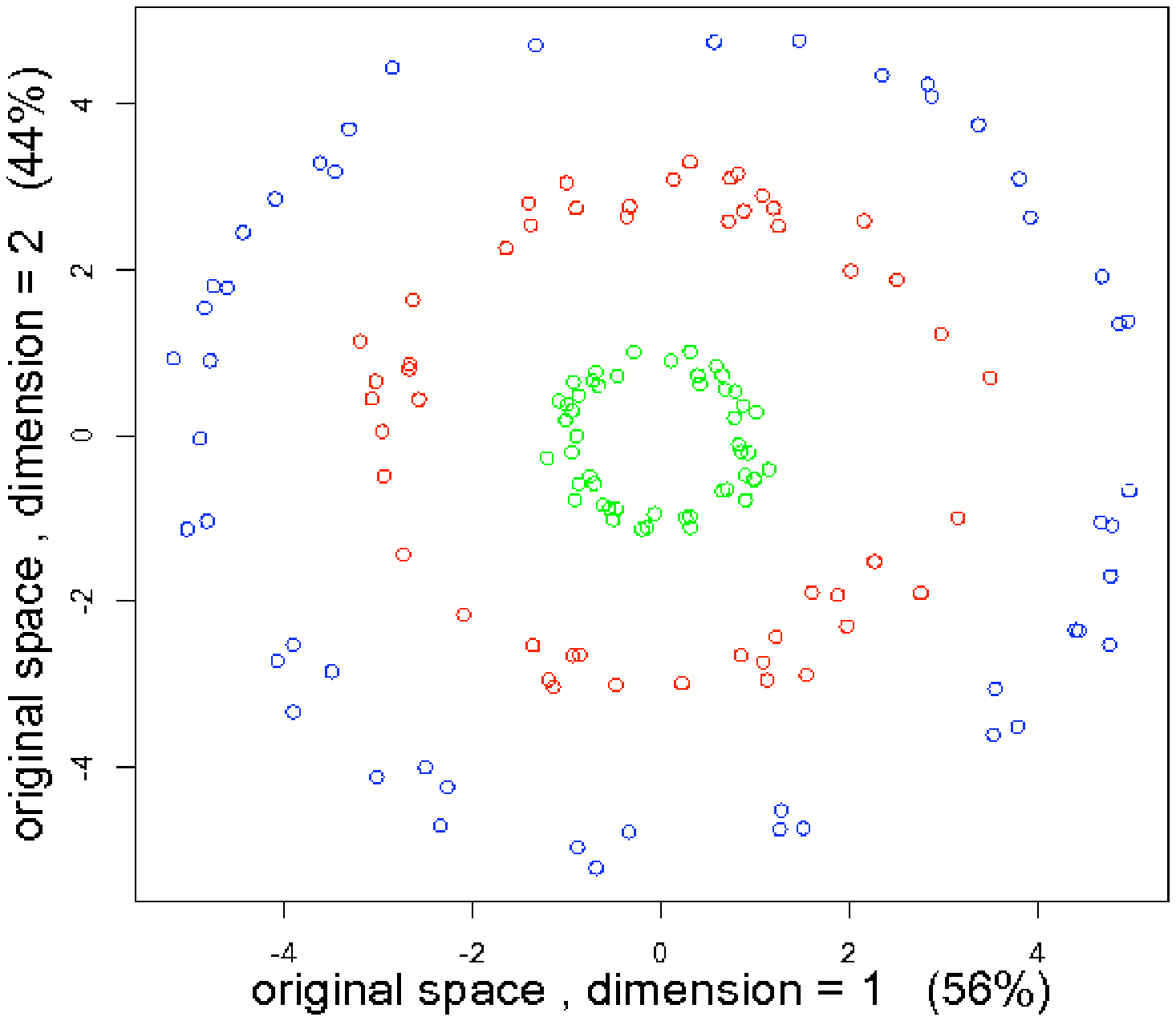}
\includegraphics[width=3.8cm]{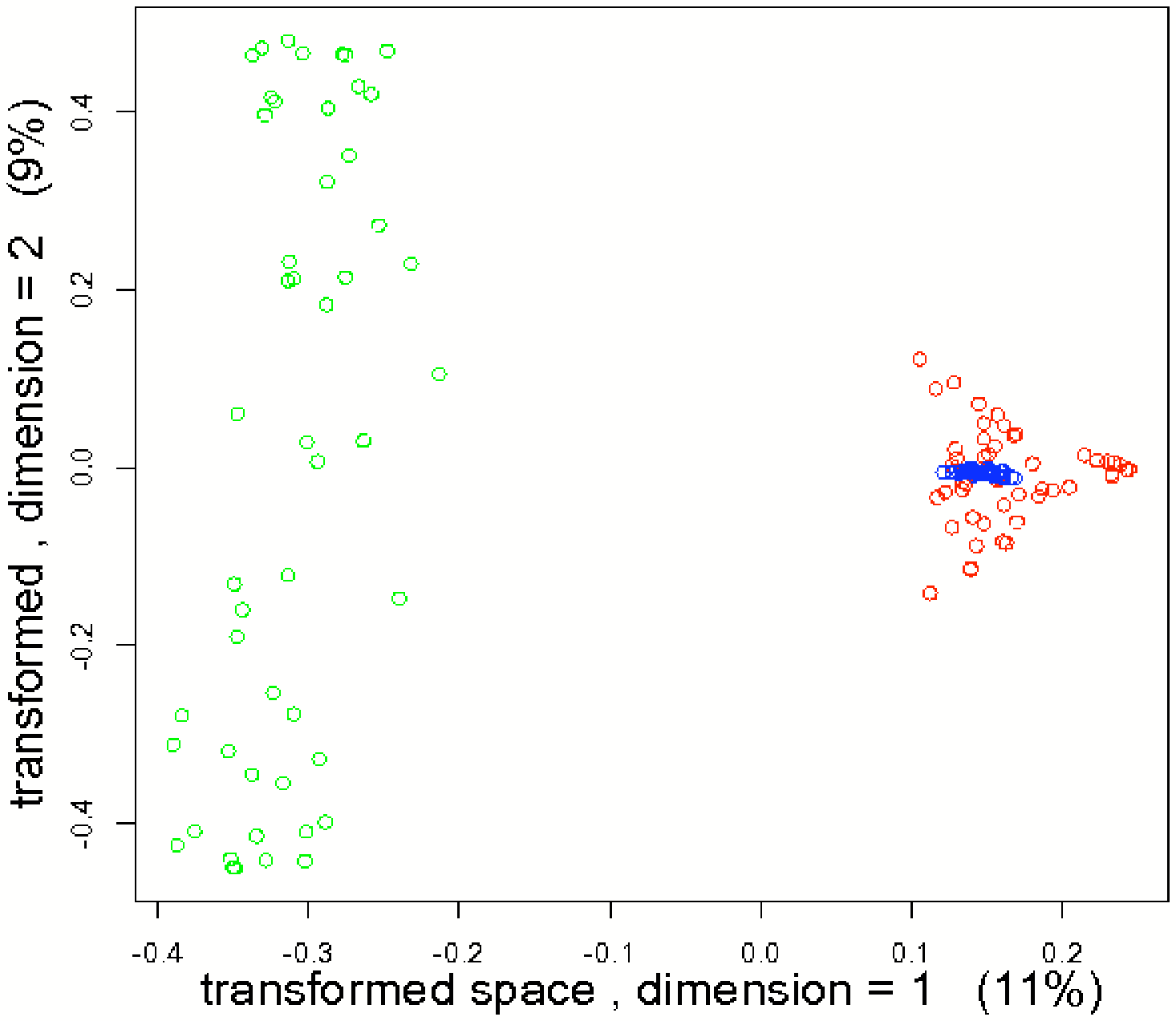}
\includegraphics[width=3.8cm]{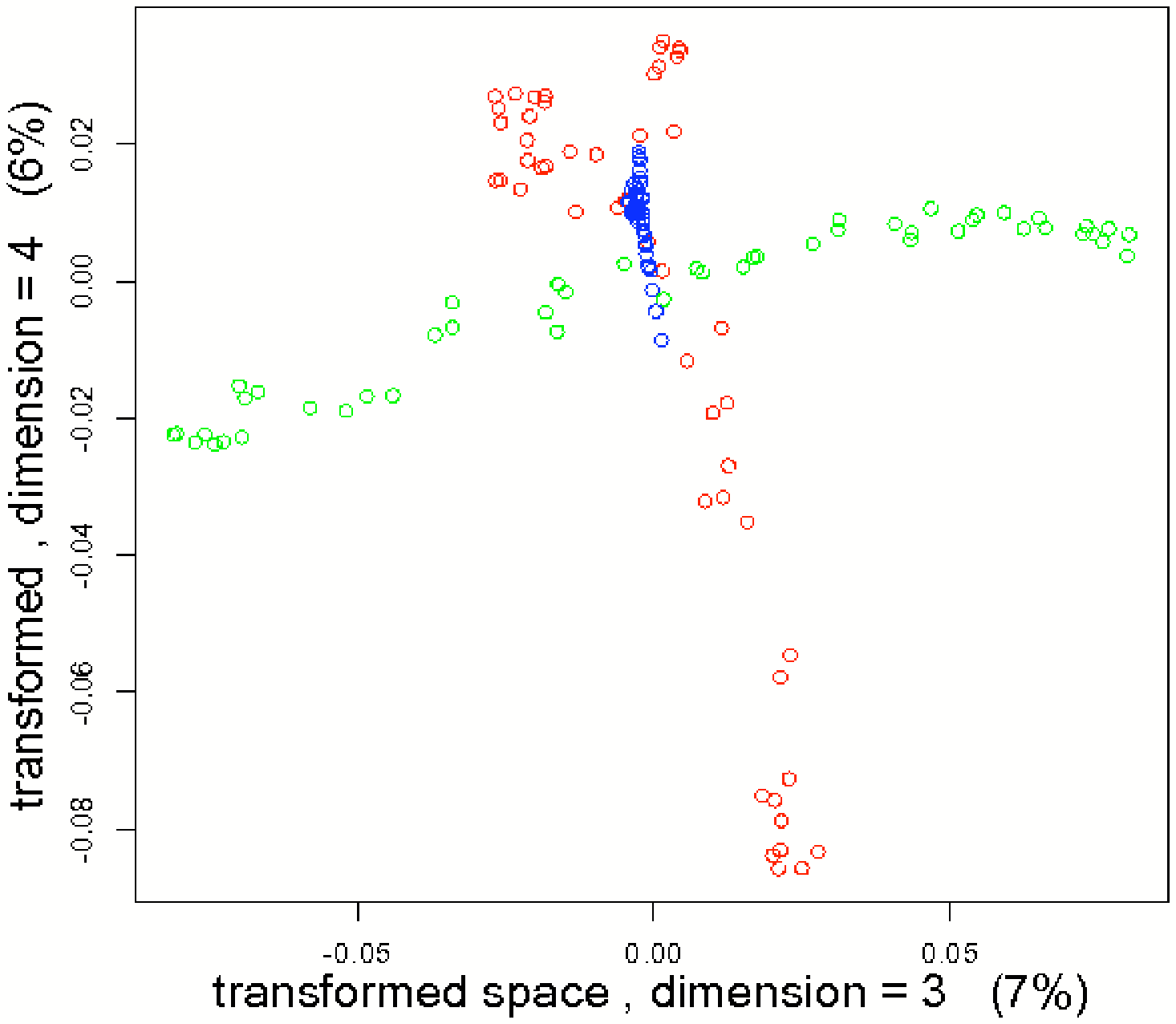}
\caption{Left: three groups of 50 individuals each, uniformly generated  on concentric circles of  radii 1, 3 and 5, with a radial standard deviation of 0.1, 0.3 and 0.2, respectively.
MDS reconstruction of the configuration transformed as $\varphi(D)=1-\exp(-0.65\:  D)$ (see text), in dimensions 1 and 2 (center) and dimensions 3 and 4 (right).}
\label{sectors}
\end{center}
\end{figure}

The same strategy is bound to fail with the data of Figure \ref{sectors} ($n=150$, 
$p=2$, $m=3$), reaching a percentage  of well-classified individuals of  35\%, close to the expected value of 33\% under random attribution. 

\begin{figure}
\begin{center}
\includegraphics[width=3.8cm]{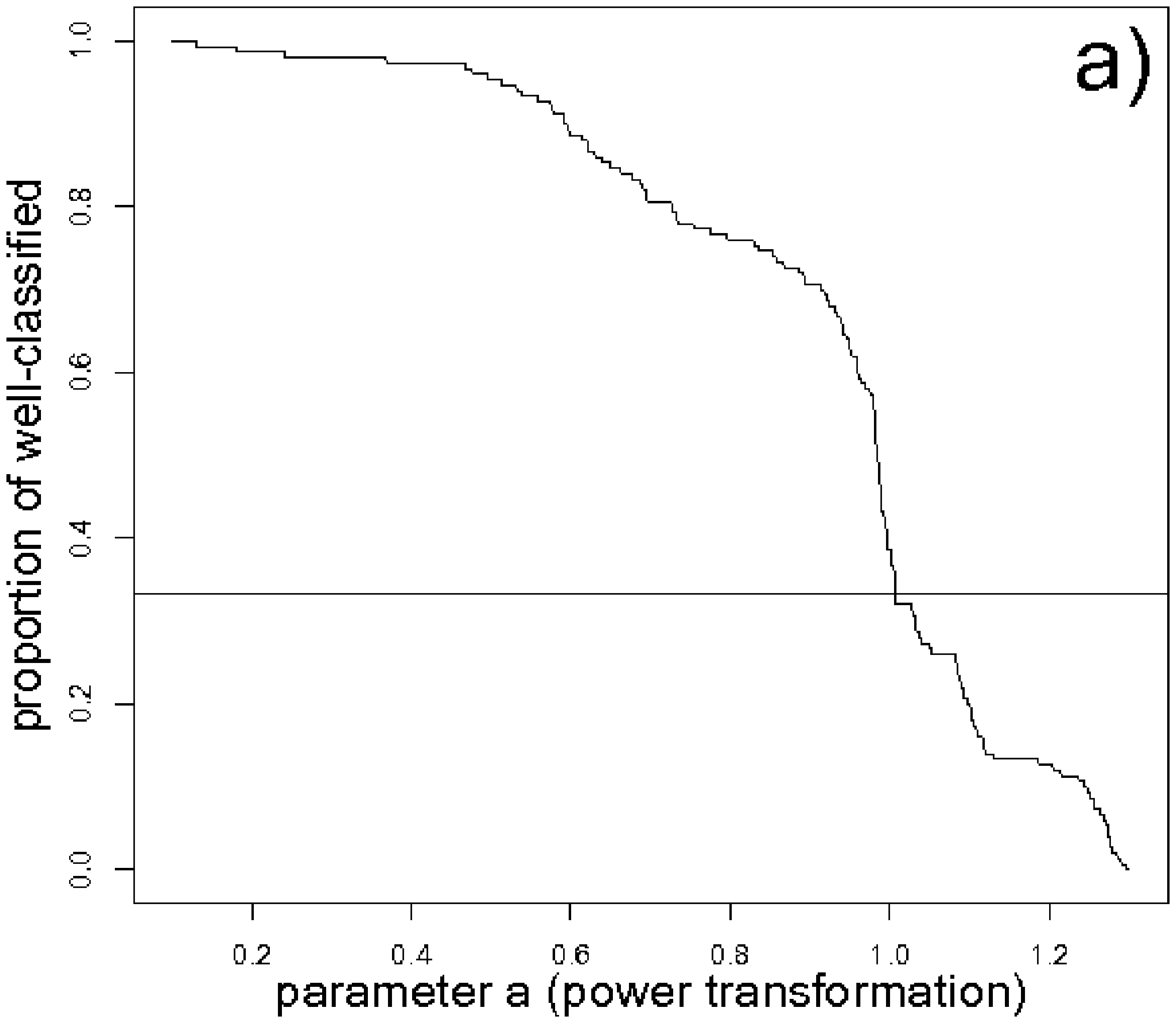}
\includegraphics[width=3.8cm]{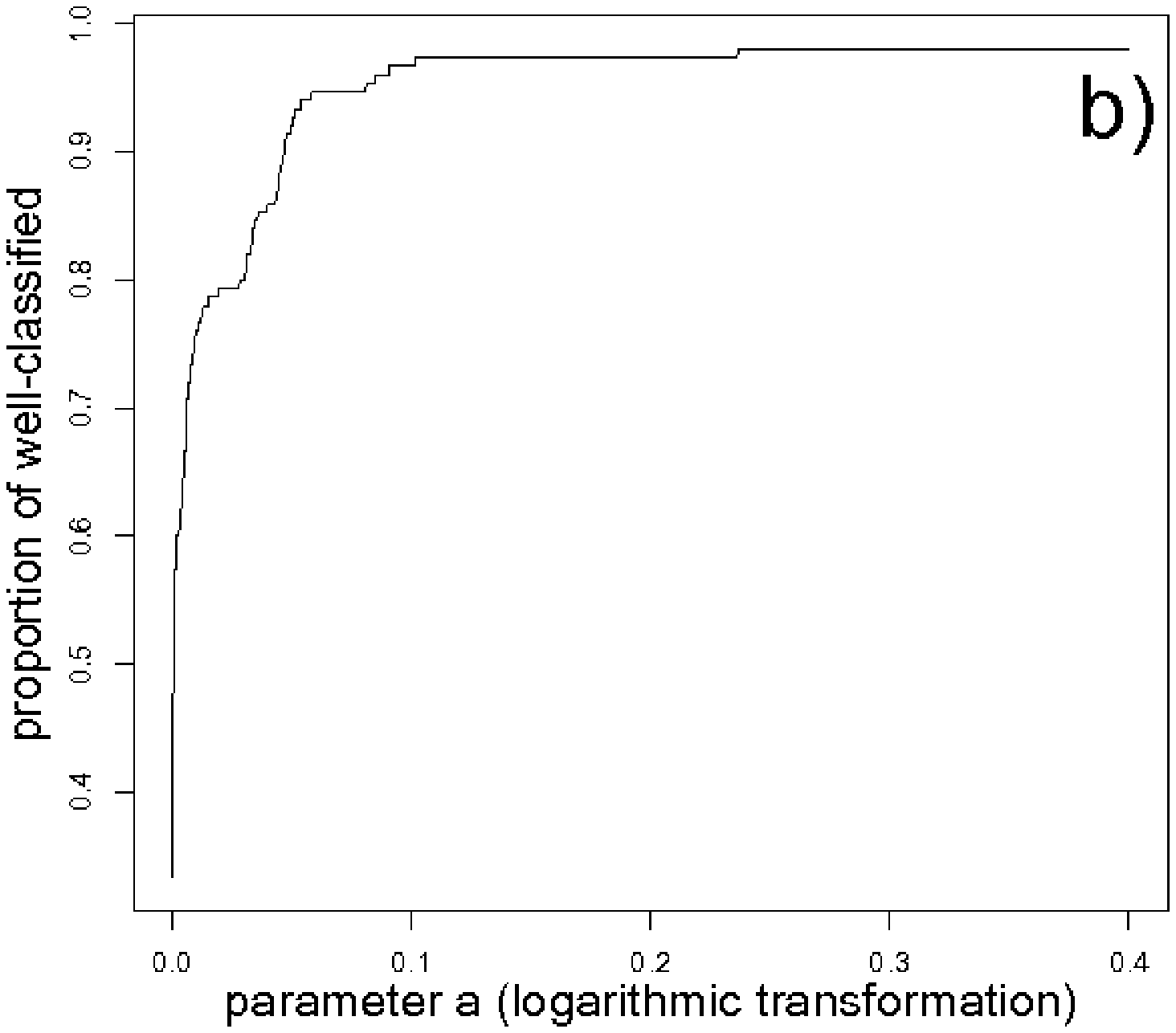}
\includegraphics[width=3.8cm]{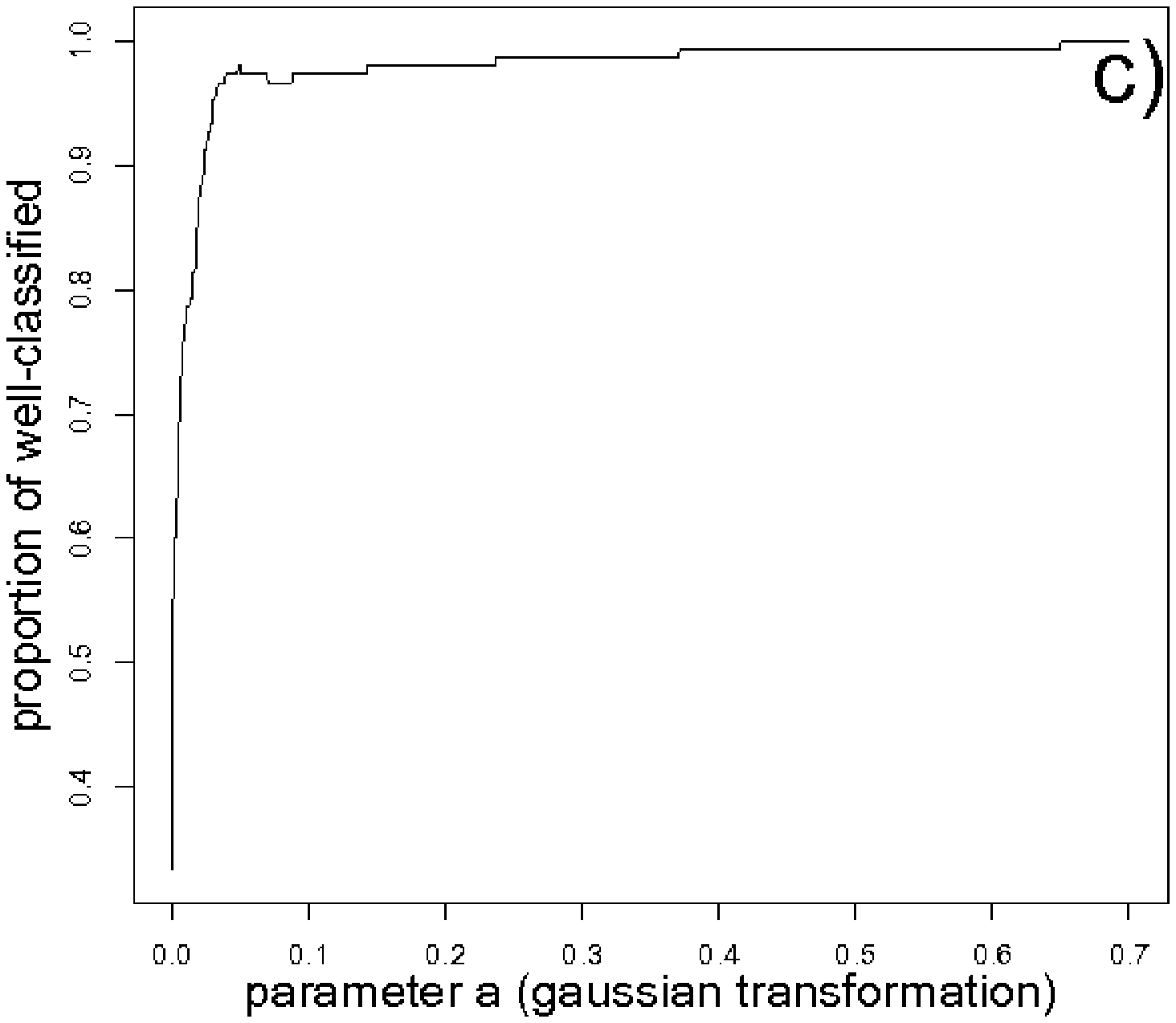}
\caption{Proportion of well-classified individuals, after Schoenberg transformation  of the original data of Figure \ref{sectors}. a) power transformation $\varphi(D)=D^{a}$; note that $a>1$ does  {\em not} corresponds to a valid transformation, and results in a {\em decrease} of the proportion below the chance level.
b) logarithmic transformation  $\varphi(D)=\ln(1+a D)$.
c) Gaussian transformation  $\varphi(D)=1-\exp(-aD)$.}
\label{propowc}
\end{center}
\end{figure}

However, linear discrimination can be attempted on  Schoenberg transformations of the original distances, resulting in the algorithm (see (\ref{huyg})): 
   
    \vspace{0.15cm}

\pagebreak 

\noindent {\tt  \underline{Distance-based discriminant algorithm}:  \\
\noindent 1) compute $\tilde{D}_{i\tilde{g}}=\sum_{j=1}^n  f^g_j \tilde{D}_{ij}-\mbox{\small $\frac12$}\sum_{j,k=1}^n f^g_jf^g_k\tilde{D}_{jk}$,\\ where $ \tilde{D}_{ij}=\varphi(D_{ij})$\\ and $f^g_j = \mbox{\small $I(i\in g)/n_g$}$ (\mbox{\small $n_g=\sum_{j\in g}1 $}) is the distribution in group $g$\\
\noindent 2) assign object $i$ to group $\arg\min_{\tilde{g}} \tilde{D}_{i\tilde{g}}$.}

 \vspace{0.15cm}
 
Figure (\ref{propowc}) shows the resulting proportion of well-classified individuals, for various one-parameter families of transformations $\varphi(D|a)$. 
In this data set, the maximum proportion of well-classified individuals reaches   100\% for the Gaussian transformation (for $a\ge0.65$). That is, a sufficiently vigorous Schoenberg transformation succeeds in mapping the initial configuration of Figure \ref{sectors} in such a way that  the three groups can be enclosed in three associated {\em disjoint} hyperspheres.

On one hand, this result is completely  expected: mapping the data into a high-dimensional feature space, in which the former become linearly separable, is a routine strategy in the Machine Learning community, developed ever since the nineties (see e.g. Chen et al. 2007 and references therein). On the other hand, the conceptual, formal and computational simplicity  of the above, presumably new algorithm, should to be emphasized.

\section{Conclusion}
\label{conc}
The Machine Learning literature contains  innumerable algorithms based upon Gaussian and other radial kernels:  the procedure exposed in Section \ref{adbda} is indeed  just one 
among many  possible applications, aimed at illustrating the operational content of the theory. Higher-order ``principled" embeddings, pioneered by the work of Vapnik (1995) and embodied in this article by the class of Schoenberg transformations, are arguably about to be 
 incorporated in standard Data Analysis, to be routinely used  in applications, and taught at   graduate and undergraduate non-specialized audiences. 

Recasting the whole  Machine Learning formalism in terms of Euclidean distances, rather than in terms of kernels, could efficiently contribute towards this assimilation: first, the statements in either formalism can be translated into the other, at granted by Theorems of Section  \ref{defthe}. In particular,  to the ``kernel trick" stating that all  the quantities of interest depend upon kernels only (and not upon the object features themselves) corresponds an equally efficient  ``distance trick", stating that Euclidean distances themselves (and not their underlying coordinates) permit to express  all the real quantities of interest, as in (\ref{huyg}), (\ref{shp}), or Section \ref{adbda}; see also Sch{\"o}lkopf  (2000) and Williams (2002). Furthermore, Euclidean distances are arguably more intuitive than kernels, as attested by the development of Geometry and Data Analysis
(including  their non-Euclidean extensions; see e.g.
Critchley and Fichet (1994) for a review). In that respect,   such a revisitation could prove itself beneficial, both from the prospect of  future scientific developments  as from a pedagogical point of view.

\end{document}